\documentclass{article}

\PassOptionsToPackage{numbers, compress, square}{natbib}

\usepackage[preprint]{neurips_2020}

\usepackage[utf8]{inputenc} 
\usepackage[T1]{fontenc}    
\usepackage{hyperref}       
\usepackage{url}            
\usepackage{booktabs}       
\usepackage{amsfonts}       
\usepackage{nicefrac}       
\usepackage{microtype}      
\usepackage{authblk}
\usepackage{soul}

\usepackage{hyperref}
\usepackage{amsmath}
\usepackage{graphicx}
\usepackage{comment}
\usepackage{multirow,bigdelim}

\usepackage{array}
\usepackage{wrapfig}

\usepackage{csvsimple}
\usepackage{adjustbox}
\usepackage[percent]{overpic}
\usepackage{makecell}

\usepackage{blindtext}

\newcolumntype{C}[1]{>{\centering\let\newline\\\arraybackslash\hspace{0pt}}m{#1}}

\newif\ifdraft
\drafttrue

\ifdraft
\newcommand{\dcc}[1]{{\color{red}[\textbf{DC:} #1]}}
\newcommand{\rgc}[1]{{\color{orange}[\textbf{RG:} #1]}}
\newcommand{\abc}[1]{{\color{blue}[\textbf{AB:} #1]}}
\newcommand{\rzc}[1]{{\color{olive}[\textbf{Rz:} #1]}}

\else
\newcommand{\dcc}[1]{}
\newcommand{\rgc}[1]{}
\newcommand{\abc}[1]{}
\newcommand{\rzc}[1]{}

\fi

\usepackage{pgfplots}
\pgfplotsset{compat=newest}
\usepgfplotslibrary{groupplots}
\usepgfplotslibrary{dateplot}

\usepackage[bottom]{footmisc}
\title{MRGAN: Multi-Rooted 3D Shape Generation with Unsupervised Part Disentanglement}

\author[1]{Rinon Gal}
\author[1]{Amit Bermano}
\author[2]{Hao Zhang}
\author[1]{Daniel Cohen-Or}

\affil[1]{Tel-Aviv University}
\affil[2]{Simon Fraser University}

\begin{document}

\maketitle

\begin{abstract}


We present MRGAN, a {\em multi-rooted\/} adversarial network which generates {\em part-disentangled\/} 3D point-cloud shapes
{\em without\/} part-based shape supervision. 
The network fuses multiple branches of tree-structured graph convolution layers which produce point clouds, with learnable constant inputs at the tree roots. Each branch learns to grow a different shape part, offering control over the shape generation at the part level. 
Our network encourages disentangled generation of semantic parts via two key ingredients: a {\em root-mixing\/} training strategy which helps decorrelate the different branches to facilitate disentanglement, and a set of loss terms designed with part disentanglement and shape semantics in mind.
Of these, a novel convexity loss incentivizes the generation of parts that are more convex, as semantic parts tend to be. In addition, a root-dropping loss further ensures that each root seeds a single part, preventing the degeneration or over-growth of the point-producing branches. We evaluate the performance of our network on a number of 3D shape classes, and offer qualitative and quantitative comparisons to previous works and baseline approaches. We demonstrate the controllability offered by our part-disentangled generation through two applications for shape modeling: part mixing and individual part variation, {\em without\/} receiving segmented shapes as input. 


\end{abstract}

\section{Introduction}
\label{sec:intro}

Generative models, and especially generative adversarial networks (GANs) \cite{goodfellow_generative_2014}, play a prominent role in modern deep learning, owing to their applicability to a multitude of tasks. In the realm of geometric deep learning, these tasks range from 3D shape editing~\cite{3DGAN,achlioptas_learning_2018} 
and completion~\cite{yang_dense_2018} to novel view synthesis~\cite{nguyen-phuoc_hologan_2019}, shape-to-shape translations~\cite{yin_logan_2019}, and even 3D scene synthesis~\cite{zhang2018_scene}.
%
However, despite the progresses made, the expressive power of GANs still leaves more to be desired. In particular, there is often a lack of
control over the attributes of the generated objects, since these attributes are all entangled in the latent space of the networks.
%
%
When considering generative modeling of 3D shapes, a promising direction is to attain controllability by learning a model with
{\em semantic part disentanglement\/}.
Part disentanglement promotes treating a shape as a {\em composable structure\/} of parts, 
rather than an inseparable whole, a view which has been shown to boost the generative diversity of shapes~\cite{schor_componet_2019}.
An intriguing question is whether a part-disentangled GAN can be trained for a shape class without segmented shapes to provide supervision. 
A true structural understanding of the shape class may very well result from such an {\em unsupervised\/} part disentanglement.

For image generation, multiple conditional models have attempted to assert control 
by injecting additional supervision into their framework~\cite{mirza_conditional_2014,sangkloy_scribbler_2016,isola_image--image_2018}, such as class labels~\cite{brock_large_2019} or segmentation masks~\cite{ferrari_generative_2018}. 
These models have managed to achieve varying degrees of disentanglement, yet they are limited by the need to acquire labeled data, which may often be prohibitively expensive, such as in the case of 3D models.
More recently, several works have shifted their attention to unsupervised disentanglement. Utilizing novel architectures designed with the task in mind, they are able to obtain control over stylistic features at varying scales~\cite{karras_style-based_2019}, pose information~\cite{nguyen-phuoc_hologan_2019}, the number of generated objects~\cite{nguyen-phuoc_blockgan_2020}, or general modalities in the data~\cite{sendik_unsupervised_2020}, without the use of any labels. All of these works, however, focus on image-space disentanglment, not part-level controls for 3D shapes.

\begin{figure}[t!]
    \centering
    \includegraphics[width=14cm, clip]{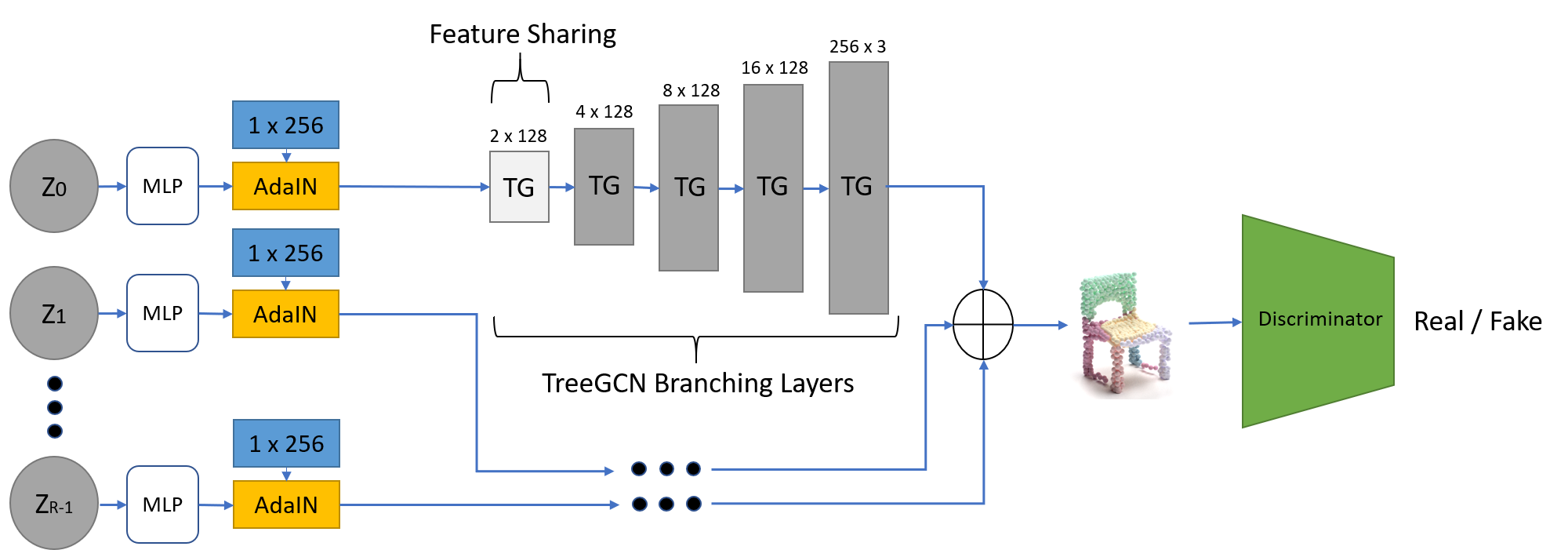}
    \caption{Network architecture of MRGAN, our multi-root 3D shape generator. For each of the $R$ roots, a latent vector is drawn from a normal distribution and applied to a learned constant (blue) via an AdaIN layer, in a similar manner to StyleGAN. The roots are grown together into a single object using TreeGCN layers, with limited feature sharing at the lowest layer.}
    \label{fig:arch-fig}
\end{figure}

In this paper, we address this gap in the realm of 3D point cloud generation, by introducing MRGAN, a {\em multi-rooted\/} adversarial network which generates {\em part-disentangled\/} 3D point-cloud shapes {\em without\/} part-based shape supervision.
Specifically, our model is split into several different generative paths, each responsible for producing a single shape part, before being assembled into a complete shape; see Figure~\ref{fig:arch-fig}. Each path begins from a learned root constant and a perturbation using a sampled latent vector, in a similar fashion to recent style based architectures such as StyleGAN~\cite{karras_style-based_2019}, and leads into a tree-structured graph convolution network which produces a point cloud. 
The roots are encouraged to grow into meaningful shape parts with no repetitions via a collection of losses. In addition to an adversarial loss,
we develop a novel {\em distance-to-hull\/} based loss by training and applying an auxiliary scoring network to encourage the generation of parts that are more convex, as semantic parts tend to be.
We further introduce a {\em root-dropping\/} loss, wherein we take the complete shape and subtract all the points emanating from a single root. Building on the intuition that a shape devoid of one of its meaningful parts (e.g., an airplane without one of the wings) is no longer recognizable as a real shape, we elect to penalize the generator if the partial shape is still recognizable as a real object by the discriminator. Lastly, we employ a triplet loss between the object parts and an identity regularization term~\cite{nguyen-phuoc_hologan_2019} applied to each path, prompting stronger localization and disentanglement of parts. 

On top of the network losses, a crucial component of our approach is a {\em root mixing\/} training strategy, where we utilize 
various combinations of latent code inputs for the different roots during training. 
This strategy not only helps decorrelate the
point-producing branches to facilitate disentanglement, but also prepares the trained MRGAN to exert control
over individual parts at inference time, as the network was already faced with various part mixing scenarios during training.

In summary, our main contributions are fourfold:
\begin{itemize}
    \item A multi-path, unconditional GAN for 3D shape generation which is, to the best of our knowledge, the first to achieve {\em unsupervised\/} part disentanglement for 3D shapes.
    \item A differentiable approach to estimating and encouraging convexity of point-cloud shapes.
    \item An intuitive point removal loss that encourages part individuality.
    \item A multi-scenario network training strategy via root mixing.
\end{itemize}

These contributions combine to form a network which learns a part-disentangled 3D shape representation.
We evaluate MRGAN on several shape classes, and compare qualitatively and quantitatively to prior works and baselines. 
The sampling of an individual latent vector for each root allows for natural control over different parts. We demonstrate such controllability through two novel applications for shape modeling: part mixing and individual part variation, {\em without\/} segmented shapes as input.
%



\section{Related work}
\label{sec:related}

\paragraph{Part-aware shape generation.}
Structure - or part-aware shape modeling has been a prominent topic in computer graphics, where new shapes are synthesized
by composing parts from a collection of {\em segmented\/} shapes~\cite{mitra_structure-aware_2013}. More recently, 
attention has been shifting towards learning generative models of 3D structures~\cite{egstar2020_struct}. In particular, the proliferation of deep learning 
has led to an extensive collection of part-based representations, ranging from boxes~\cite{li_sig17_GRASS,tulsiani_learning_2018}, convexes~\cite{chen_BSP_2020, deng_CVX_2020}, and super quadratics~\cite{paschalidou_superquadrics_2019}, to implicit indicator functions~\cite{chen_BAE_2019} and deformable meshes~\cite{gao_siga19_SDM}.

For 3D shape generation, some recent works~\cite{dubrovina_composite_2019,schor_componet_2019} have resorted to 
Spatial Transformer Networks~\cite{jaderberg_spatial_2016}, where others have adopted recursive neural networks~\cite{li_sig17_GRASS} or
graph neural networks~\cite{mo2019structurenet}. The generative models employed include GANs~\cite{wang_global--local_2019}, variational autoencoders (VAEs)~\cite{schor_componet_2019,gao_siga19_SDM}, or a combination of the two~\cite{li_sig17_GRASS}, while the networks were all trained with collections of segmented shapes or structural shape representations such as part hierarchies~\cite{li_sig17_GRASS} or graphs~\cite{mo2019structurenet}.




The key difference between all of these prior works and ours is that we learn to generate part-disentangled 3D shapes in an unsupervised manner. Our network is trained with no supervision at the part level, nor any segmented shapes or part collections to draw from. Furthermore, while we use a multi-path approach, our network is trained as a single uniform block, intended to create a fully cohesive shape from the onset. 
Unsupervised learning of semantic parts often must rely on auxiliary part priors such as compactness~\cite{chen_BAE_2019} and convexity~\cite{vankaick14convexseg}.
In a recent work, BSP-Net, \citet{chen_BSP_2020} obtain unsupervised part-aware shape reconstruction using a union of convexes assembled from a set of binary space partitions. However, their network is an implicit decoder and does not generate shapes like a GAN or claim individual control over the generation of the convexes.


\vspace{-5pt}

\paragraph{Unsupervised disentanglement.}
The learning of disentangled generative models aims to gain control over individual factors of variation in the generated samples while leaving the other factors unchanged~\cite{locatello_challenging_2019}. 
%
Broadly speaking, unsupervised disentanglement is the task of learning such controls without any prior knowledge regarding the factors of variation we wish to disentangle. While recent work suggests that such solutions are indeed fundamentally impossible without using some inductive bias on the model or the data~\cite{locatello_challenging_2019}, we are content with the commonly approached subset of the task --- learning a disentangled representation without the use of explicit, labeled data. Multiple works have tackled this task~\cite{higgins_beta-vae_2017,kim_disentangling_2019}, offering ways to disentangle specific properties~\cite{nguyen-phuoc_hologan_2019, nguyen-phuoc_blockgan_2020} or sets of properties divided by scale~\cite{karras_style-based_2019}. Some works are designed to encourage control over a pre-determined property, such as object pose~\cite{nguyen-phuoc_hologan_2019}, while others are more general, proposing control over modalities in the data~\cite{sendik_unsupervised_2020} without prior knowledge of what form these modalities might take.

Our method falls firmly into the camp of networks designed for disentangling 
over pre-determined, yet latent, properties, i.e., the geometric and structural properties of shape parts. These are clearly distinctive from scale-dependent properties such as the stylistic features learned by StyleGAN~\cite{karras_style-based_2019}. Furthermore, whereas many of the prior works address only global properties of the resulting objects, e.g., pose and class identity, our method allows for control over more localized traits.

\vspace{-5pt}

\paragraph{Neural Point Cloud Generation.}
Compared to 3D shape representations such as meshes and voxels, there has been relatively thin literature on neural generation of point clouds, perhaps owning to their irregular and unordered nature. The earliest fully generative models have appeared as recently as \citet{achlioptas_learning_2018}. Their approach, however, had difficulty in exploiting localized features~\cite{valsesia_learning_2019}, prompting more recent works to consider hierarchical sampling operations~\cite{pointflow, li2018point} and graph convolutions~\cite{shu_3d_2019, valsesia_learning_2019} which allows for the same, while maintaining permutational invariance. Unlike initial graph convolutional methods~\cite{kipf_semi-supervised_2017}, these generative works had to contend with a non-constant graph structure, with deeper layers containing more vertices and a different connectivity structure. Their solutions involved dynamic generation of adjacency matrices~\cite{valsesia_learning_2019} or a tree-structured approach in the case of TreeGAN~\cite{shu_3d_2019}, where graph connectivity lies between points and their ancestors in previous layers, rather than their same-layer neighbours.
Both of these graph-based works displayed a natural {\em clustering\/} of the generated points, 
with TreeGAN even demonstrating a connection between the shared ancestry of points and their relative spatial positions in the final cloud.

Our work takes this natural clustering a step further by directly encouraging the clusters to take the shape of meaningful parts. Most importantly, whereas these learned models are fully entangled, with no control over the individual point clusters, we split the network structure into multiple, independently controlled trees. This allows us to achieve more than just a natural clustering, but a disentangled representation with individual control over the shape and number of generated parts.

\section{Method}
\label{sec:method}

Given a set of objects belonging to a single class, our goal is to learn a latent representation and a generator that can map vectors sampled from this representation into novel instances of the object class. Furthermore, we want the latent representation to be disentangled in parts, such that we can modify only a single part of the generated object, but leave the remainder untouched.
We achieve this disentanglement via a forest-like set of tree-convolution paths, as well as a set of network losses and training strategies, designed to encourage separation with no explicit part labels.

\subsection{Multi-Rooted Tree-Convolutional GAN}

Inspired by the natural ability of graph convolutional networks (GCNs) to generate 3D point clouds composed of localized clusters~\cite{shu_3d_2019}, we elect to employ such a network - TreeGAN - as the backbone of our generator.
%
The standard graph convolutional approach, first described by \cite{kipf_semi-supervised_2017}, considers the entire graph at each layer, combining features of a static number of points with fixed connectivity. The tree-structured approach instead considers points according to their {\em depth\/} in a tree. 

\vspace{-5pt}

\paragraph{Tree-GCN for point cloud generation.}
Our tree begins with a root and progressively grows into a set of vertices through the application of convolution and branching operations. These operations increase the depth of the tree by creating a set of child nodes for each leaf in the current tree. Each layer, then, is responsible for extending the depth of the tree and calculating the features of the newly created leaves. 
In this approach, connections to neighbours in a graph are replaced with connections to a node's ancestors in the tree, such that feature propagation is given by:
%
\begin{equation}
    v_{i}^{l + 1} = \sigma\left(W^{l}_{i,i}v_{i} + \sum_{v_{j} \in A\left(v_{i}\right)}{W^{l}_{i, j}v_{j}} + b^{l}\right),
\end{equation}
where $v_{i}$ is the i-th node in the graph, $A\left(v\right)$ are the ancestors of node $v$, $v^{l}$ denotes the set of features of node $v$ at layer $l$, $W^{l}$ and $b^{l}$ are learned weights and bias terms for the l-th layer, and $\sigma$ is an activation function.
{\em Branching\/} is performed by an additional multiplication:
\begin{equation}
    v_{i,j}^{l + 1} = B_{i,j}^{l+1} \dot v_{i}^{l + 1},
\end{equation}
where $v_{i,j}^{l + 1}$ is the j-th child of vertex i at layer l + 1, and $B_{i,j}^{l+1}$ is a learned branching-weight term.

This framework lends itself well to the traditional generative approach of increased resolution at deeper layers, and the common ancestors have been shown to drive points on the same branch to display similar properties (e.g., have similar spatial coordinates). However, it does not allow for individual control over these clusters, nor does it make an effort to ensure that the clusters are meaningful or avoid overlaps between the clusters.

\vspace{-5pt}

\paragraph{Multi-rooted GAN.}
To encourage the latent space to display the desired part disentanglement, we make the separation of parts {\em explicit\/} in our generator architecture, breaking the generative path into {\em multiple\/} parallel networks with limited information sharing.

As shown in Figure \ref{fig:arch-fig}, we begin with $R$ root constants, each of dimension $1 \times 256$. For each root, we sample a latent vector $z \in \mathcal{N}\left(0, 1\right)^{96}$, pass it through an MLP network, and apply the output to the root through an AdaIN layer, following the same concepts shown in StyleGAN~\cite{karras_style-based_2019}.
Each root is then fed as the initial point into the TreeGCN framework, growing into an independent part.

To ensure that the resulting object maintains coherence, e.g., the different parts border each other at the right places, we allow some information to propagate between parts at the lowest layer, via {\em feature sharing\/}. The feature vector of each point is concatenated with a vector of features max-pooled over all trees and passed through an MLP in order to reduce it back to the original dimension. 



The points for each part are given by the leaves of the lowest level of each tree. After generating all the individual parts, their points are concatenated, resulting in a point cloud as the network output.

For a discriminator, we follow the architecture of \cite{achlioptas_learning_2018}. The full architecture details as well as a complete illustration of the feature sharing layers are provided in the supplementary material.

\subsection{Network losses}

While our multi-path architecture offers a clear way to influence each tree separately, there is no guarantee that the trees will converge into unique, meaningful parts. Indeed, applying only a discriminative loss to the generated cloud results in poor localization and redundancy between parts. In order to encourage a meaningful separation, we apply a set of additional losses:



\begin{wrapfigure}{R}{0.45\textwidth}
    \centering
    \vspace{-10pt}
    \includegraphics[width=0.35\textwidth]{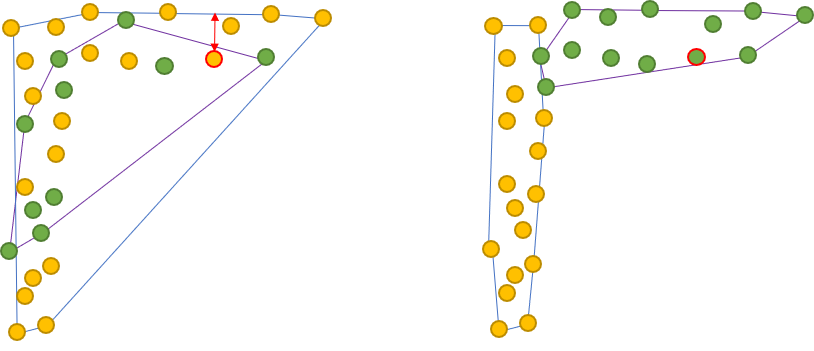}
    \caption{Illustration of the principle behind our convexity loss. Consider points representing the seat 
    and a leg 
    of a chair that are split in two ways: randomly on the left and semantically on the right. 
    On the left, points in neither class exhibit convexity, hence they tend to lie further away from the boundary of the convex hull of their class.}
    \label{fig:convex-example}
    \vspace{-10pt}
\end{wrapfigure}

\textbf{Convexity loss.} We draw inspiration from previous works that eschew the traditional semantic partitioning, but choose instead to consider an approach that recognizes a shape as composed from multiple approximately convex parts~\cite{kaick_shape_2014,gadelha_label-efficient_2020, chen_BSP_2020}. While not strictly semantic, these weakly convex parts are still strongly associated with semantic parts~\cite{vankaick14convexseg,deng_CVX_2020}.

However, known measures for weak convexity are generally not differentiable or highly expensive to compute. We side-step both issues by defining a simple {\em differentiable\/} measure that correlates with convexity. Given a cloud of points, we measure the {\em average distance\/} from the points to the nearest hyperplane of the convex hull of the points.
It is expected that a point cloud which encompasses a convex part will have relatively more of its points on or near that hull, leading to a relatively small {\em distance-to-hull\/}, while the opposite is true for a cloud encompassing several parts; see Figure~\ref{fig:convex-example} for a contrast.

We train an auxiliary network to predict the distance-to-hull measure for a given point cloud. The network follows the same form as the discriminator, and uses an $L_2$ metric as the optimization objective. Complete architecture and training details are provided in the supplementary material.

The convexity loss for the generator is given by applying the distance-to-hull prediction network to the points generated from each root separately, and taking the mean over the results: 
$L_{h} = \frac{1}{R} \sum_{i=1}^{R}F\left(G_{i}\left(z\right)\right),$
where $F(p)$ is the result of applying the auxiliary network to a set of points and $G_{i}\left(z\right)$ is the set of points generated from the tree originating at the $i$-th root.

\textbf{Root dropping loss.}
An unfortunate side effect of the global operations used to side-step the cloud permutation concerns is that our discriminator becomes less sensitive to repetitions amongst different parts. So long as a good approximation of the object surface can be achieved, parts may find themselves overlapping partially or entirely with each other. As our aim is for the parts to become distinct, we discourage this behaviour via a root-dropping loss.

This loss is calculated by taking the difference of the discriminator's score on the full shape and the score returned for the shape without the points originating at the dropped root, normalized by the full-shape score, as per:
\begin{equation}
    L_{rd} = \frac{1}{R}\sum_{i=1}^{R}\frac{D\left(G(z)\right) - D\left(G(z)\backslash G_{i}(z)\right)}{D\left(G(z)\right)},
\end{equation}
where $D(p)$ is the discriminator score for a set of points and $G(z)\backslash G_{i}(z)$ is the set of generated points excluding those originating at the $i$-th root.

If the trees display considerable overlap, the discriminator returns a similar score with and without the repeated root. Lack of repeats, meanwhile, results in the discriminator being significantly more confident that the part-less shape is a fake, and thus the loss is decreased.

\textbf{Triplet and reconstruction losses.}
While neither is unique to our work, we employ two additional losses.
The first is a triplet loss \cite{schroff_facenet_2015}, employed between generated points, where their source root index is used as a class label. This loss is intended to encourage the points to better localize. We denote it as $L_{t}$.
Lastly, we employ an identity regularization term similar to the one proposed by HoloGAN~\cite{nguyen-phuoc_hologan_2019}. This loss is calculated by appending an additional head to the discriminator, one that is meant to reconstruct the latent vectors that were initially provided to the network. The loss term is the $L_{2}$ distance between the original sampled latent codes and their reconstruction. The aim of this loss is to discourage the network from ignoring any of the inputs and instead abusing the feature-sharing layer to dictate their appearance. We denote this loss by $L_{rec}$.

In total, our loss term is given by:
\begin{equation}
    L_{total} = L_{WGAN-GP} + \lambda_{h}\cdot L_{h} + \lambda_{rd}\cdot L_{rd} + \lambda_{t}\cdot L_{t} + \lambda_{rec}\cdot L_{rec} ,
\end{equation}
where we used $\lambda_{h} = \lambda_{t} = \lambda_{rec} = 1.0$ and $\lambda_{rd} = 0.1$ for our experiments.

\subsection{Network Training and Root-Mixing}


\vspace{-5pt}

\paragraph{Root-mixing training strategy.}
%
%


We add additional regularization to our network in the form of a root-mixing training strategy, where we utilize different combinations of latent code inputs for the different network roots, according to the following uniformly sampled strategies:

\begin{itemize}
    \item Sampling a single $z\in\mathcal{N}\left(0, 1\right)^{96}$ and feeding it into all roots.
    \item Sampling two latent vectors $z_1, z_2 \in \mathcal{N}\left(0, 1\right)^{96}$, feeding $z_1$ into a random half of the roots and $z_2$ to into the other half.
    \item Sampling two latent vectors $z_1, z_2 \in \mathcal{N}\left(0, 1\right)^{96}$, feeding $z_1$ into a randomly chosen root and $z_2$ into the rest.
    \item Sampling a random latent code for each root.
\end{itemize}

In this we draw from the lessons of StyleGAN, where \citet{karras_style-based_2019} show that adding a degree of latent code mixing between different scales during the training process assists in improving disentanglement and the quality of generated outputs. However, their best results are achieved when employing mixing only half the time. We stipulate that a degree of non-mixed inputs helps the network stability and convergence as the network need deal only with fewer factors of variation. We have designed our strategy to reflect that.

\vspace{-5pt}

\paragraph{Training and implementation details.}
We trained our network on the ShapeNet \cite{shapenet2015} dataset, making use of 3 different object classes: chair, table and airplane.
For the chair data set we allow the network to make use of 6 roots, while for both the table and airplane data sets we make use of 5 roots.

For all data sets we train the network for 500 epochs using an Adam Optimizer with a learning rate of $1e-4$, $\beta_1 =0$ and $\beta_2=0.99$. We perform 8 optimization steps on the discriminator for every step of the generator. Our latent codes were sampled from a normal distribution $\mathcal{N}\left(0, 1\right)$. Our branching operations were of sizes: [2, 2, 2, 2, 16] leading to a total of 256 points generated from each root.

\section{Experiments}
\label{sec:exp}

\begin{figure}[!t]
    \centering
    \includegraphics[width=0.99\linewidth]{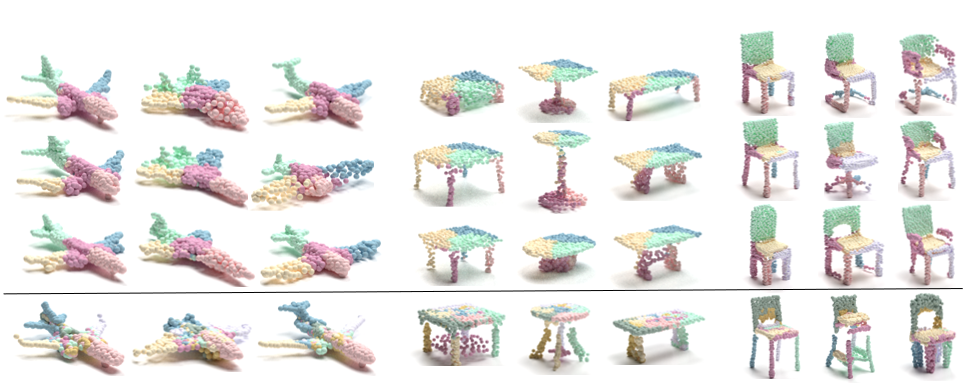}
    \caption{Generated samples from the airplane, table, and chair classes. The last row was generated using a vanilla TreeGAN \cite{shu_3d_2019} implementation and is provided for comparison. For each class, all points originating in a single root share the same color. We use 5 roots for the airplane and table classes and 6 roots for the chair class. As TreeGAN is grown from a single root, we colored the points according to their ancestor at the 8-point level.}
    \label{fig:mix-samples}
\end{figure}

\begin{figure}[!t]
    \centering
    \includegraphics[width=0.95\linewidth]{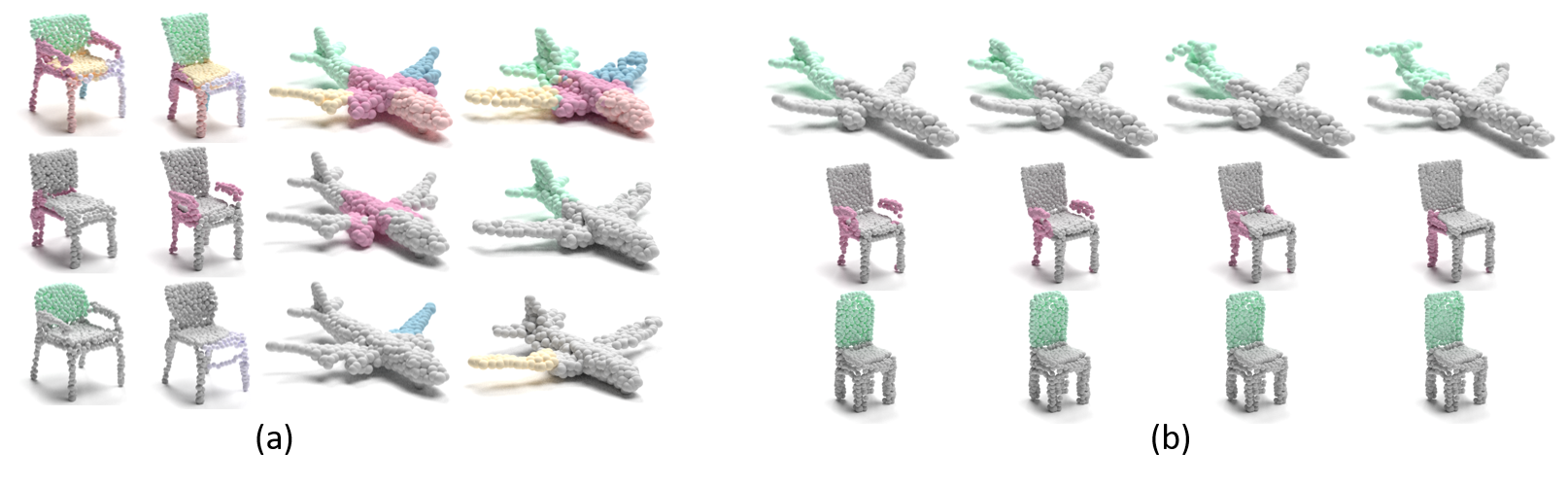}
    \caption{Single root modification experiment results. (a) Top row: original shapes, parts colored by path of origin. Bottom rows: a single (colored) root is randomly changed, while keeping all others unaffected 
    (grey). (b) Interpolation of individual part latent codes. For each row we smoothly modify the latent code of the colored part, and leave the remainder unchanged (grey). }
    \label{fig:mix-single}
\end{figure}

\begin{figure}[!t]
    \centering
    \includegraphics[width=0.99\linewidth]{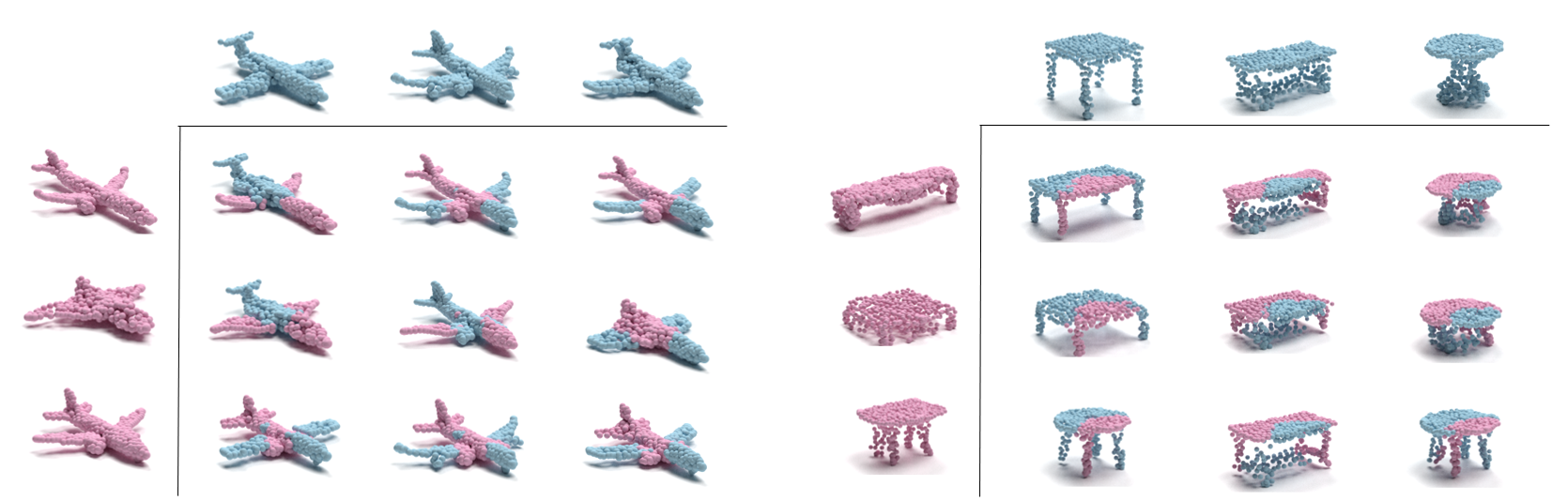}
    \caption{Part mixing. For each class, the shapes in the left column and the top row were generated from a single latent vector each. The rest of the shapes were generated by utilizing the latent vector from the top row for half of the roots, and the latent vector from the left column for the rest.}
    \label{fig:mix-multi}
\end{figure}

We demonstrate the benefits of MRGAN's multi-rooted approach with a set of experiments. Figure \ref{fig:mix-samples} shows a collection of generated shapes from the airplane, table and chair classes, with points colored according to the path they belong to. Note the comparison to TreeGAN's generated shapes, whose parts are not so cleanly divided between the branches. In Figure \ref{fig:mix-single} we show the results of editing a generated shape by changing the latent code provided to one root at a time, including part-code interpolation. In Figure \ref{fig:mix-multi} we demonstrate the mixing of latent codes between two shapes from the same object class. By utilizing the latent codes from one shape for half the roots and the latent codes from the other for the remainder, we are able to generate a new shape that combines traits from both.

We evaluate the quality of disentanglement in our network by modifying the latent code for a single part and measuring the change it undergoes when compared to the rest of the points. Simple distance metrics are ill suited for our needs due to the minor global changes that our generated shapes undergo in order to preserve coherence. As such, we propose to measure instead the fraction of points in each part that undergo a \textit{meaningful} change, where we consider a change to be meaningful if it is greater than the mean movement of points between pairs of randomly sampled shapes, calculated over 2500 pairs. We also use this metric to conduct an ablation study of our proposed losses and mixing techniques. The results are provided in Table \ref{ablation-table}. Our ablation experiments resulted in three key observations: the reconstruction loss is of considerable importance in maintaining disentanglement, and any attempt to discard the feature sharing layer or the root-mixing training strategy lead to the network failing to converge or to generated shapes losing coherence, depending on what other losses are present. The benefits of our remaining losses are best appreciated visually. We provide a qualitative overview of their effects in the supplementary materials.

Lastly, we compare our results to previous cloud generation networks using the metrics of \cite{achlioptas_learning_2018}. Coverage results are shown in Table \ref{cmpr-table}. The full comparison can be found in the supplementary materials. These coverage results demonstrate that our network is capable of generating more diverse shapes. This mirrors the observations of \citet{schor_componet_2019}, which observe that part-based models can lead to an increase in the diversity of generated content.

\begin{table}[]\centering

\begin{tabular}{|l|ccc|}
\hline \bf Model & \bf \thead{Modified part} & \bf \thead{Unmodified parts} & \bf \thead{Ratio}\\ \hline
Chair & 0.096 & 0.017 & 5.36 \\
Table & 0.080 & 0.029 & 2.71  \\
Airplane & 0.122 & 0.026 & 4.66  \\
\hline
Airplane - Mixing + GAN loss & 0.107 & 0.050 & 2.13 \\
+ Hull-distance loss & 0.109 & 0.048 & 2.23 \\
+ Triplet loss & 0.123 & 0.050 & 2.46 \\
+ Reconstruction loss & 0.122 & 0.026 & 4.66 \\ 
\hline 
\end{tabular}\newline



\caption{The fraction of points that undergo a meaningful change when modifying a single part, for both points within the modified part and outside it. A larger ratio between the two scores indicates improved disentanglement. Results without any mixing or feature sharing are not provided as they do not generate coherent shapes. }\label{ablation-table} 
\end{table}

\begin{table}[t]\centering
\begin{tabular}{|l|l|cccccc|}
\hline \bf \multirow{2}[2]{1.5cm}{Class} & \bf \multirow{2}[2]{1.5cm}{Metric} & \multicolumn{2}{c}{\bf \thead{r-GAN}} & \multicolumn{2}{c}{\bf \thead{Valsesia et al}} & \bf \multirow{2}[2]{1.5cm}{\thead{tree-GAN}} & \bf \thead{MRGAN} \\ 
& & dense & conv & no up. & up. & & \bf (ours)\\\hline
\multirow{2}[2]{1.5cm}{Chair} & \bf \thead{COV-CD} & 33 & 23 & 26 & 30 & {58} & \bf{67} \\
 & \bf \thead{COV-EMD} & 13 & 4 & 20 & {26} & \bf{30} & 23 \\

\hline
\multirow{2}[2]{1.5cm}{Airplane} & \bf \thead{COV-CD} & 31 & 26 & 24 & 31 & {61} & \bf{75} \\
 & \bf \thead{COV-EMD} & 9 & 7 & 13 & 14 & 20 & \bf{21} \\
\hline
\multirow{2}[2]{1.5cm}{Table} & \bf \thead{COV-CD} & - & - & - & - & 71 & \bf{78} \\
& \bf \thead{COV-EMD} & - & - & - & - & \bf{48} & 31 \\
\hline 

\end{tabular}\newline
\caption{Coverage metric comparisons to previous generative point cloud models, using the metrics of \cite{achlioptas_learning_2018}. We use the values reported by \cite{shu_3d_2019} where applicable. The best results for each metric are in bold. Previous works did not provide evaluation metrics on the table class. For the sake of completeness, we provide coverage metrics for tables using our own network and a tree-GAN model trained by us.}\label{cmpr-table}
\end{table}

\section{Conclusions}

We presented MRGAN, a multi-rooted architecture that addresses the challenging task of learning a part-disentangled representation in an unsupervised manner. Our work shows that, as in the image domain, providing the network with a strong inductive bias can help overcome the need for detailed labeling. This in turn allows us to train on unlabeled sets and still achieve part-level modifications at inference time.
In order for the network to achieve disentanglement at test time, a degree of root, i.e., part-level mixing is also crucial at training time. This mirrors the results of StyleGAN \cite{karras_style-based_2019}, where employing mixing through different scales was shown to assist in attribute separability.

We emphasize that the part-level editing shown in Figures~\ref{fig:mix-single} and~\ref{fig:mix-multi} differs from conventional part assembly~\cite{mitra_structure-aware_2013} in two ways. First, part assembly always operates on {\em pre-segmented\/} shapes, while our editing/mixing inputs latent codes, never shape parts. Second, while the shape parts can undergo a variety of deformations during part assembly, such deformations are typically performed 
only to achieve {\em local\/} part alignment and fitting.
In contrast, our network arrives at a {\em globally coherent\/} and well-connected shape in an end-to-end manner. As the different roots are
allowed to affect each other, a point cloud part grown from the same latent code may develop into different shapes according to {\em non-local\/} 
contexts and these shape changes are not confined to any pre-defined deformation space --- this is especially apparent in the first row of the table-mixing example in Figure~\ref{fig:mix-multi}.

The multi-path architecture offers further advantages which we have yet to touch upon. With the parts borne from different paths, a natural extension could be to allow different paths to generate a different number of points, helping overcome the drop in resolution for parts that cover a large volume. Taking the idea a step further, hierarchical representations may be explored, where any natural symmetry or other geometric relations between parts could be directly exploited.

\clearpage

\section*{Broader Impact}

Our work tackles the generation and part-level editing of shapes represented by point-clouds. Its architectural concepts and novel losses could be employed in the context of other deep-learning-based tasks, however we do not believe this work otherwise poses any noticeable ethical or social implications.

\medskip

\small

\bibliographystyle{plainnat}
\bibliography{egbib}

\end{document}


\maketitle
\tableofcontents
\clearpage

\appendix
\section{Additional results}

\begin{figure}[H]
    \centering
    \includegraphics[width=1.0\linewidth]{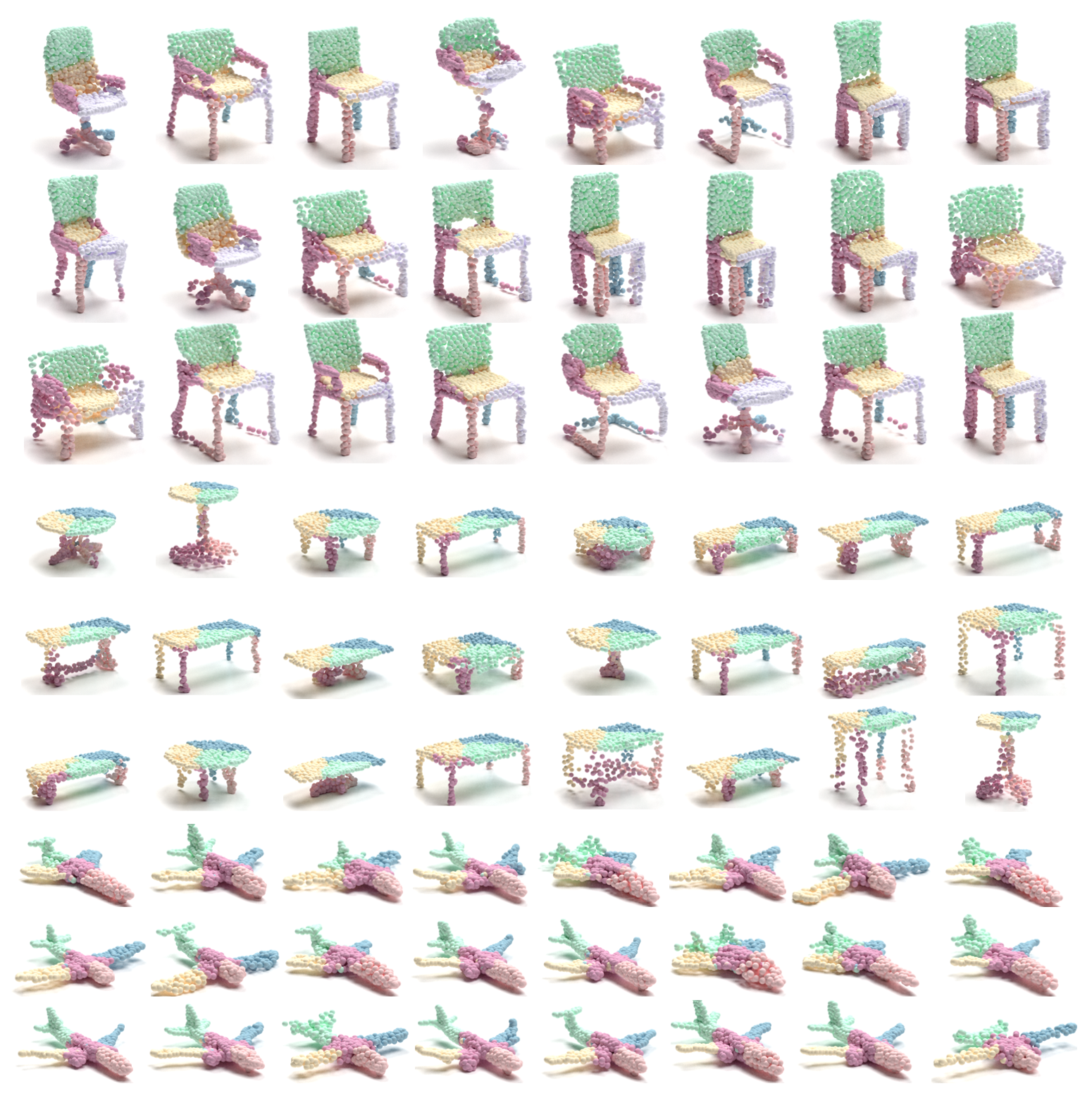}
    \caption{Additional samples generated for the chair, table and airplane classes. Points are colored according to the root they originate from. We used 6 roots for the chairs and 5 roots for the tables and airplanes.}
    \label{fig:more-samp}
\end{figure}

\begin{figure}[H]
    \centering
    \includegraphics[width=0.5\linewidth]{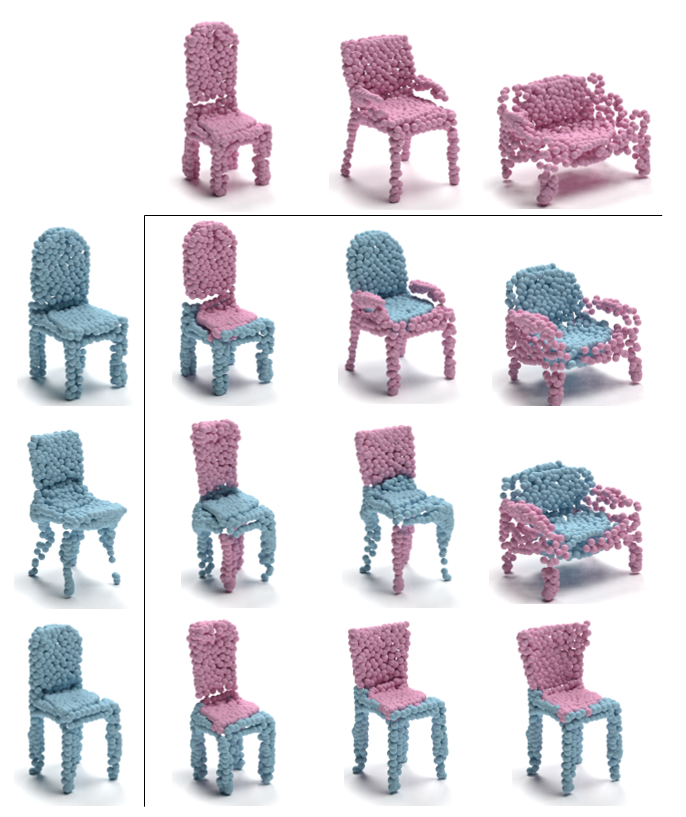}
    \caption{Part mixing results for the chair class. The shapes in the left column and the top row were generated from a single latent vector each. The rest of the shapes were generated by utilizing the latent vector from the top row for some of the roots, and the latent vector from the left column for the rest.}
    \label{fig:chair-mix}
\end{figure}

In Figure \ref{fig:more-samp} we show additional samples generated from our three classes. In Figure \ref{fig:chair-mix} we provide samples of shape mixing for the chair class. 

Note in particular that part mixing does not simply consist of copying the parts as-is, but that the network attempts to maintain their identity while preserving the overall coherence. This is particularly prominent in the second row of Figure \ref{fig:chair-mix}, where the round chair back is maintained, but its size and inclination are adjusted to fit with the rest of the chair.

\begin{figure}[H]
    \centering
    \includegraphics[width=0.99\linewidth]{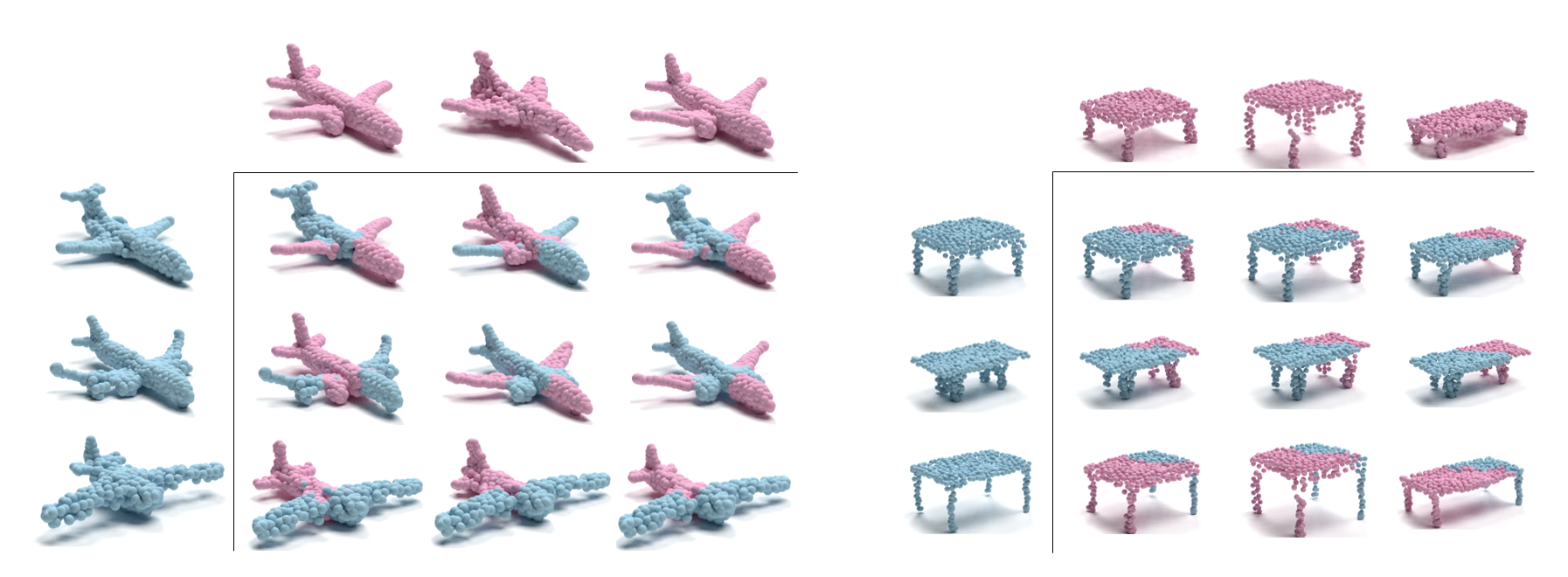}
    \caption{Additional part mixing results for the airplane and table classes. For each class, the shapes in the left column and the top row were generated from a single latent vector each. The rest of the shapes were generated by utilizing the latent vector from the top row for some of the roots, and the latent vector from the left column for the rest.}
    \label{fig:air-table-mix}
\end{figure}

Figure \ref{fig:air-table-mix} shows additional mixing results for the airplane and table classes. The coherence preserving effect can be observed in the table class. The identity of legs (e.g. their inclination or curvature) is maintained, but the height and width of the resulting table are adjusted in order to avoid an imbalance.

\section{Comparisons on additional metrics}

We provide our comparisons results on all metrics of \cite{achlioptas_learning_2018} in Table \ref{cmpr-table-supp}.
While our quality metrics do not compete with the state of the art, they remain within the same order or improve upon some of the previous works.

\begin{table}[H]\centering
\begin{tabular}{|l|l|ccccc|}
\hline \bf Class & \bf Model & \bf\thead{JSD} & \bf \thead{MMD-CD} & \bf \thead{MMD-EMD} & \bf \thead{COV-CD} & \bf \thead{COV-EMD} \\ \hline
& r-GAN (dense) & 0.238 & 0.0029 & 0.136 & 33 & 13 \\
& r-GAN (conv) & 0.517 & 0.0030 & 0.223 & 23 & 4  \\
\multirow{1}[3]{1.5cm}{Chair} & Valsesia et al. (no up.) & \color{blue}{0.119} & 0.0033 & 0.104 & 26 & 20  \\
 & Valsesia et al. (up.) & \color{red}{0.100} & 0.0029 & \color{red}{0.097} & 30 & \color{blue}{26}  \\
  & tree-GAN & \color{blue}{0.119} & \color{red}{0.0016} & \color{blue}{0.101} & \color{blue}{58} & \color{red}{30}  \\
    & MRGAN (ours) & 0.246 & \color{blue}{0.0021} & 0.166 & \color{red}{67} & 23  \\
\hline
& r-GAN (dense) & 0.182 & 0.0009 & 0.094 & 31 & 9 \\
& r-GAN (conv) & 0.350 & 0.0008 & 0.101 & 26 & 7  \\
\multirow{1}[3]{1.5cm}{Airplane} & Valsesia et al. (no up.) & 0.164 & 0.0010 & 0.102 & 24 & 13  \\
 & Valsesia et al. (up.) & \color{red}{0.083} & 0.0008 & \color{blue}{0.071} & 31 & 14  \\
  & tree-GAN & \color{blue}{0.097} & \color{red}{0.0004} & \color{red}{0.068} & \color{blue}{61} & \color{blue}{20}  \\
    & MRGAN (ours) & 0.243 & \color{blue}{0.0006} & 0.114 & \color{red}{75} & \color{red}{21}  \\
\hline
\multirow{1}[3]{1.5cm}{Table} & tree-GAN & 0.077 & 0.0018 & 0.082 & 71 & 48 \\
& MRGAN (ours) & 0.287 & 0.0020 & 0.155 & 78 & 31  \\
\hline 

\end{tabular}\\
\caption{Comparisons to previous generative point cloud models on the metrics of \cite{achlioptas_learning_2018}. We use the values reported by \cite{shu_3d_2019} where applicable. For the airplane and chair classes, the best and second best results are colored in red and blue respectively.}\label{cmpr-table-supp}
\end{table}

\section{Qualitative ablation}
\begin{figure}[H]
    \centering
    \includegraphics[]{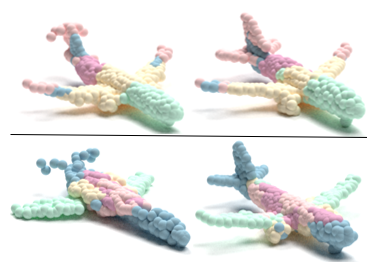}
    \caption{Loss removal effects. Top row: the root dropping loss was removed, leading to a redundancy between the orange and blue roots. Bottom row: the convexity loss was removed, leading to considerable degradation in separation into meaningful parts.}
    \label{fig:miss-loss}
\end{figure}

The effects of some of our losses are better appreciated visually. In Figure \ref{fig:miss-loss} we provide a visual contrast of the their effects. Removal of the root-dropping loss (top row) results in the emergence of redundant parts (blue and orange roots). Removal of the convexity loss (bottom row) results in considerable degradation in the separation to meaningful parts. 

\section{Modification locality visualization}
\begin{figure}[H]
    \centering
    \includegraphics[width=0.99\linewidth]{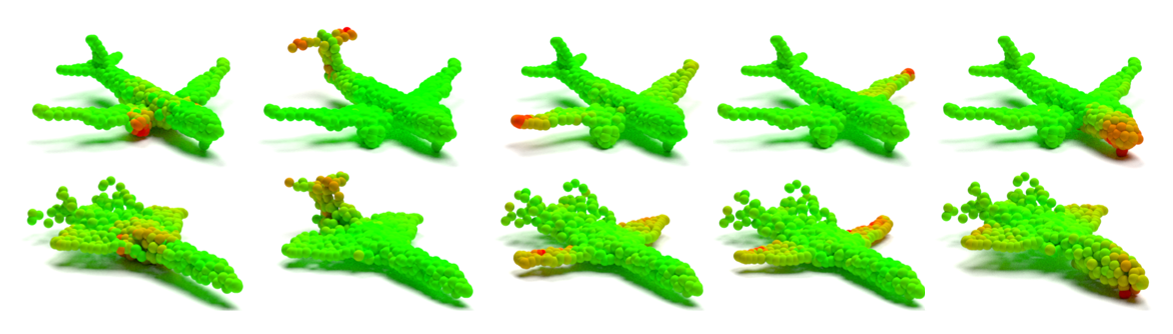}
    \caption{A visualization of root modification locality. Each image displays a heat map of individual point coordinate changes. Red indicates the point that had the largest shift while green indicates the smallest (or no) shift.}
    \label{fig:dist-map}
\end{figure}

We provide an additional visualization of the disentanglement afforded by the network via a distance-based mapping (Figure \ref{fig:dist-map}). In each image, we modify a single root and color all points in the cloud according to their distance from their location in the source shape. Red indicates a greater shift in the coordinates of a point, while green indicates a smaller or no change.

This distance-based visualization offers us another indication that the results of modifying a single root are largely constrained to a single part. Some changes lead to a more global shift. In particular, the bottom right airplane had an overall translation upwards due of the addition of a wheel to the nose. Despite this overall shift, the identities of the other parts are largely unchanged.

\section{Feature sharing layer}

The feature sharing layer is implemented by a series of operations. We first max-pool all feature vectors over the point dimension. The resulting maximal feature vector is passed through a dense layer and then concatenated back to the feature vector of each individual point. Finally, the features of each point are passed through a 1D convolutional operator (with weights shared between the different points), using a kernel size of 1 and a number of filters equal to the original dimension of point features.

\begin{figure}[H]
    \centering
    \includegraphics[width=0.99\linewidth]{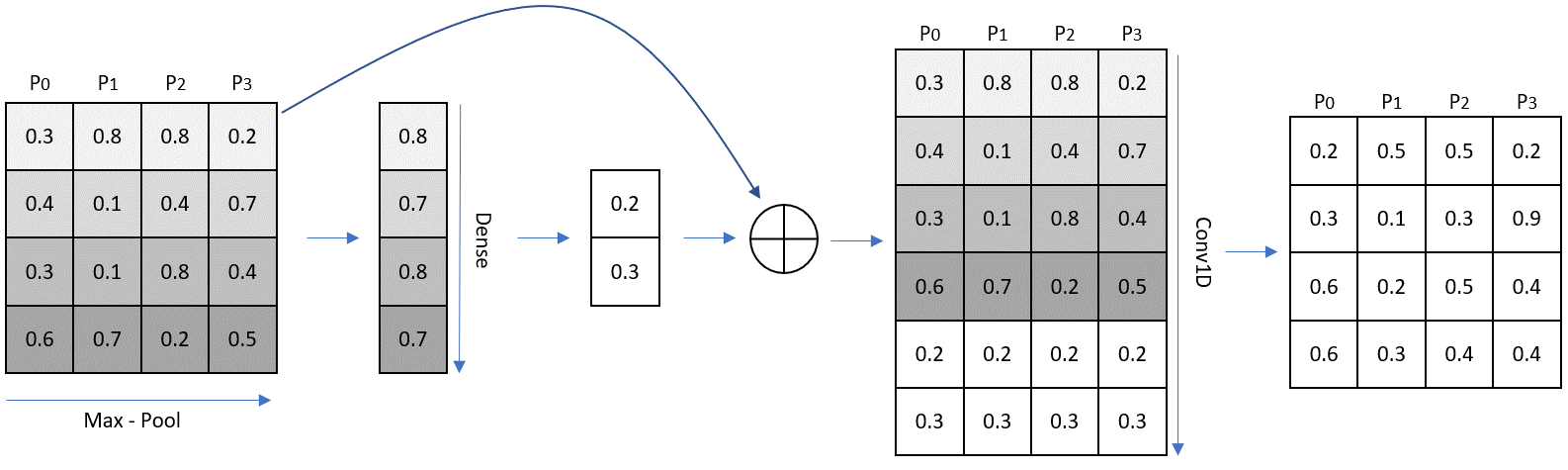}
    \caption{A visualization of the steps undertaken in the feature sharing layer.}
    \label{fig:feat-share}
\end{figure}

A visualization of this process is provided in Figure \ref{fig:feat-share}. The layer dimensions are provided in Table \ref{tab:feat-share}.

\begin{table}[H]
    \centering
    \begin{tabular}{|l|cccc|}\hline
        Layer type & Kernel size & Stride & Activation & Output dimension  \\\hline
        MaxPool1D & P & 1 & - & 1 x F \\
        Dense & - & - & - & 1 x 16 \\
        Concat & - & - & - & P x (F + 16) \\
        Conv1D & 1 & 1 & LeakyReLU & P x F \\\hline
    \end{tabular}\\
    \caption{Network architecture of a feature sharing layer with P input points and F input features.}
    \label{tab:feat-share}
\end{table}

\section{Network architecture}
We describe the full network architecture for a generator with R roots in Table \ref{tab:gen-arch}. The discriminator architecture is provided in Table \ref{tab:disc-arch}.
Table \ref{tab:recon-arch} provides the architecture details for the additional head used to reconstruct the initial latent codes from the discriminator features.

\begin{table}[H]
    \centering
    \begin{tabular}{|l|ccc|}\hline
        Layer type & Branching ratio & Activation & Output dimension  \\\hline
        Learned root constants & - & -  & R x 256 \\
        TreeGCN  & {2} & - &  2R x 128 \\
        Feature sharing & - & LeakyReLU & 2R x 128 \\ 
        TreeGCN & {2} & - &  4R x 128 \\
        TreeGCN & {2} & - &  8R x 128 \\
        TreeGCN & {2} & - &  16R x 128 \\
        TreeGCN & {16} & LeakyReLU & 256R x 3 \\\hline
    \end{tabular}
    \caption{Network architecture of the multi-rooted generator with R roots. The root constants are normalized via an AdaIN layer, using a different latent vector as input for each root.}
    \label{tab:gen-arch}
\end{table}

\begin{table}[H]
    \centering
    \begin{tabular}{|l|cccc|}\hline
        Layer type & Kernel size & Stride & Activation  & Output dimension  \\\hline
        Conv1D & 1 & 1 & LeakyReLU &   256R x 3 \\
        Conv1D & 1 & 1 & LeakyReLU &   256R x 64 \\
        Conv1D & 1 & 1 & LeakyReLU &   256R x 128 \\
        Conv1D & 1 & 1 & LeakyReLU &   256R x 512 \\
        Conv1D & 1 & 1 & LeakyReLU &   256R x 1024 \\
        MaxPool1D & 256R & 1 & -  &  1 x 1024 \\
        Dense & - & - & - & 1 x 1024 \\
        Dense & - & - & - & 1 x 512 \\
        Dense & - & - & - & 1 x 512 \\
        Dense & - & - & - & 1 x 1 \\\hline
    \end{tabular}
    \caption{Network architecture of the discriminator, adapted from \cite{achlioptas_learning_2018}.}
    \label{tab:disc-arch}
\end{table}

\begin{table}[H]
    \centering
    \begin{tabular}{|l|cccc|}\hline
        Layer type & Kernel size & Stride & Activation & Output dimension  \\\hline
        Discriminator-Conv & - & - & LeakyReLU & 256R x 1024 \\
        MaxPool1D & 256 & 256 & - & R x 1024 \\
        Conv1D & 1 & 1 & LeakyReLU & R x 512 \\
        Conv1D & 1 & 1 & LeakyReLU & R x 128 \\
        Conv1D & 1 & 1 & LeakyReLU & R x 128 \\
        Conv1D & 1 & 1 & Tanh & R x 96 \\\hline
    \end{tabular}
    \caption{Network architecture of the discriminator's reconstruction head, used for the identity regularization term. The network begins from the outputs of the last convolutional layer of the discriminator.}
    \label{tab:recon-arch}
\end{table}

\section{Hull-distance network details}
The network architecture for our auxiliary distance-to-hull prediction network is given in Table \ref{tab:aux-arch}. For pooling, we employ both a maximum pooling and a minimum pooling operation over the point dimension, and concatenate their results along the feature axis.

\begin{table}[H]
    \centering
    \begin{tabular}{|l|cccc|}\hline
        Layer type & Kernel size & Stride & Activation  & Output dimension  \\\hline
        Conv1D & 1 & 1 & LeakyReLU &   N x 3 \\
        Conv1D & 1 & 1 & LeakyReLU &   N x 64 \\
        Conv1D & 1 & 1 & LeakyReLU &   N x 128 \\
        Conv1D & 1 & 1 & LeakyReLU &   N x 256 \\
        Conv1D & 1 & 1 & LeakyReLU &   N x 512 \\
        Pooling & N & 1 & - & 1 x 1024 \\
        Dense & - & - & LeakyReLU & 1 x 512 \\
        Dense & - & - & LeakyReLU & 1 x 256 \\
        Dense & - & - & LeakyReLU & 1 x 128 \\
        Dense & - & - & LeakyReLU & 1 x 64 \\
        Dense & - & - & LeakyReLU & 1 x 31 \\
        Dense & - & - & LeakyReLU & 1 x 1 \\\hline
    \end{tabular}
    \caption{Network architecture of the auxiliary hull-distance prediction network. The pooling operation is the concatenation of both a min-pool and a max-pool operation employed along the point dimension. }
    \label{tab:aux-arch}
\end{table}

We train the network on a combination of sparse point samples from the ShapeNet \cite{shapenet2015} training set (including all available classes) as well as synthetic points sampled from a sphere, a box, or any combination of the two with randomly sampled positions.

For each sampled cloud, we calculate the convex hull and compute the distance-to-hull metric analytically. The network is trained to predict this value for the input cloud, using an $L_2$ distance as the loss.

We employ an Adam optimizer with a learning rate of 0.001, $\beta_1 = 0.9$ and $\beta_2 = 0.999$.

The network is trained over 20,000 batches of 64 clouds each. In order to match the number of points in our generator's paths, we sample 256 points per cloud.

\bibliographystyle{plainnat}
\bibliography{egbib}